\documentclass[conference]{IEEEtran}
\IEEEoverridecommandlockouts

\usepackage[T1]{fontenc}
\usepackage{cite}
\usepackage{amsmath,amssymb,amsfonts}
\usepackage{algorithm}
\usepackage{algpseudocode}
\usepackage{graphicx}
\usepackage{textcomp}
\usepackage{xcolor}
\usepackage{booktabs}
\usepackage{multirow}
\usepackage{array}
\usepackage{listings}
\usepackage{tikz}
\usetikzlibrary{arrows.meta,positioning,fit,backgrounds,calc}
\usepackage{pgfplots}
\pgfplotsset{compat=1.17}
\usepackage[hidelinks]{hyperref}
\usepackage{url}
\usepackage{relsize}

\definecolor{ucblue}{RGB}{31,78,121}
\definecolor{uclight}{RGB}{222,235,247}
\definecolor{ucgray}{RGB}{110,110,110}
\definecolor{ucorange}{RGB}{204,102,0}

\lstdefinestyle{yaml}{
  basicstyle=\ttfamily\scriptsize,
  columns=fullflexible,
  keepspaces=true,
  frame=tb,
  framerule=0.4pt,
  showstringspaces=false,
  commentstyle=\color{ucgray}\itshape,
  morecomment=[l]{\#},
  stringstyle=\color{ucblue},
  morestring=[b]",
}
\lstdefinestyle{py}{
  language=Python,
  basicstyle=\ttfamily\scriptsize,
  columns=fullflexible,
  keepspaces=true,
  frame=tb,
  framerule=0.4pt,
  showstringspaces=false,
  keywordstyle=\color{ucblue}\bfseries,
  commentstyle=\color{ucgray}\itshape,
  stringstyle=\color{ucorange},
}

\newcommand{\uc}{\textsc{Undercurrent}}
\newcommand{\code}[1]{\texttt{\smaller #1}}

\begin{document}

\title{Undercurrent: Production-Grade Activation Probing\\for Live LLM Inference}

\author{\IEEEauthorblockN{Alizishaan Khatri}
\IEEEauthorblockA{\textit{Wrynx Inc.}\\
research@wrynx.com}}

\maketitle

\begin{abstract}
Probes trained on the internal activations of large language models (LLMs) can
detect properties such as deception, harmful intent and toxicity, often before
any problematic text has been emitted. Existing tooling for reading
activations, however, is built for offline interpretability research and does
not run inside the serving engines that host production traffic. We present
\uc{}, an open-source (Apache-2.0) Python library that runs activation probes
on a model's internal state \emph{while it generates}. A declarative YAML
specification selects which tensors to capture, at which layers and token
positions, and which probe receives them. Each extraction point runs either
\emph{inline}, on the generation path where a probe can abort the request
before the next token, or \emph{asynchronously}, on bounded per-request queues
with explicit overflow policies so observation adds no latency to the
response. Inline probes may be bounded by a timeout with fail-open or
fail-closed fallbacks and a circuit breaker. \uc{} ships adapters for Hugging
Face Transformers and for vLLM, where a worker extension maps the rows of each
continuously batched forward pass back to per-request token positions. We
describe the design, the vLLM integration and its correctness checks, and
report measurements from the library's own benchmark harness. \uc{} is
available at \url{https://github.com/wrynx/undercurrent} and on PyPI.
\end{abstract}


\section{Introduction}
\label{sec:intro}

A growing body of work shows that the hidden states of transformer language
models encode properties of the computation that are not visible in the
output text. Linear classifiers on intermediate activations, or
\emph{probes}~\cite{alain2016probes,belinkov2022probing}, recover whether a
model ``believes'' a statement to be true~\cite{burns2023latent,azaria2023lying,marks2023geometry},
expose high-level concepts such as honesty and power-seeking~\cite{zou2023repe},
and can detect backdoored ``sleeper agent'' behaviour from a single direction in
the residual stream~\cite{macdiarmid2024sleeper}. Because these signals are
present while the model is still processing its input, a probe can in
principle flag a generation \emph{before} the problematic tokens are produced,
which a text-level classifier such as Llama Guard~\cite{inan2023llamaguard}
cannot do.

Moving such probes from a notebook into production exposes a tooling gap.
Interpretability libraries such as TransformerLens~\cite{nanda2022transformerlens},
pyvene~\cite{wu2024pyvene} and baukit~\cite{bau2022baukit} give researchers
rich access to model internals, but they run models in-process, one experiment
at a time, and have no notion of a serving request, a latency budget or a
failure policy. Production LLMs, in contrast, are served by engines such as
vLLM~\cite{kwon2023vllm} that use continuous batching~\cite{yu2022orca}: a
single forward pass interleaves prefill chunks and decode tokens from many
requests, and the mapping from tensor rows to requests changes on every step.
A team that wants to run a trained probe on live traffic is therefore left to
fork the serving code, write per-architecture hooks, and solve request
attribution, concurrency, backpressure and fault isolation on its own.

\uc{} fills this gap. It is a small, layered library whose responsibility ends
where the probe begins: it captures the activations a specification asks for,
delivers them to per-request probe instances with well-defined ordering and
isolation guarantees, and turns probe verdicts into actions (abort) or
observations (sinks and metrics) under explicit latency contracts. It does not
ship trained probes; users bring their own.

This paper makes the following contributions:
\begin{itemize}
  \item A \textbf{declarative extraction-point specification} (YAML with a JSON
  Schema) and a position-selector grammar that address single tokens or
  continuous generated ranges without model surgery (Sec.~\ref{sec:spec}).
  \item An \textbf{execution model} that separates \emph{inline} probes, which
  can abort generation, from \emph{asynchronous} probes, which observe through
  bounded per-request queues with selectable overflow policies; inline probes
  can carry timeouts, fail-open/fail-closed fallbacks and a circuit breaker
  (Sec.~\ref{sec:exec}).
  \item \textbf{Engine adapters} for Hugging Face Transformers and vLLM,
  including a step-wise row-to-position mapper that makes activation capture
  correct under continuous batching and chunked prefill (Sec.~\ref{sec:adapters}).
  \item An \textbf{open-source implementation} with an overhead benchmark
  harness, a cross-backend equivalence test, and an end-to-end probe-training
  example, together with measurements and an honest account of current
  limitations (Secs.~\ref{sec:impl}--\ref{sec:limits}).
\end{itemize}

\section{Background and Related Work}
\label{sec:related}

\subsection{Probing model internals}
Probing classifiers were introduced as a diagnostic for what intermediate
representations encode~\cite{alain2016probes}; subsequent work examined their
pitfalls, such as probes learning the task themselves rather than reading it
from the representation~\cite{hewitt2019control,belinkov2022probing}. In
transformer LMs the \emph{residual stream}, the running sum to which every
attention and MLP block writes~\cite{elhage2021framework}, has become the
standard probing site. Probes on residual-stream activations detect truthfulness~\cite{burns2023latent,azaria2023lying,marks2023geometry,li2023iti},
broader concepts~\cite{zou2023repe} and deceptive behaviour~\cite{macdiarmid2024sleeper}.
A related line of work \emph{edits} activations to steer behaviour~\cite{turner2023actadd,li2023iti}
or to disrupt harmful representations~\cite{zou2024circuitbreakers}. \uc{} is
complementary to this research: it is the deployment substrate for a probe
once one has been trained and evaluated.

\subsection{Interpretability tooling}
TransformerLens~\cite{nanda2022transformerlens} re-implements many
architectures with named hook points and is a common starting point for
circuit analysis. pyvene~\cite{wu2024pyvene} expresses interventions as
serialisable configurations. baukit~\cite{bau2022baukit} provides lightweight
\code{Trace} contexts over named modules. NNsight~\cite{fiottokaufman2025nnsight}
builds intervention graphs from Python tracing code and, with NDIF, runs them
against remotely hosted models; it also offers a vLLM backend. These tools
optimise for expressiveness during research: arbitrary reads and writes,
patching and attribution. None of them provides a request-level abort, a
latency contract for probe execution, or out-of-band observation with
backpressure, which are the concerns \uc{} addresses.

\subsection{Guardrails}
Text-level guardrails operate on model inputs and outputs. Llama
Guard~\cite{inan2023llamaguard} is a fine-tuned LLM that classifies prompts and
responses against a hazard taxonomy; NeMo Guardrails~\cite{rebedea2023nemo}
wraps LLM calls with programmable input, dialogue and output rails. Both work
with any model, including ones behind an API, but they see only text, and
streaming output rails check text in chunks. Activation probes use different
evidence and can act per token; the two approaches are complementary, and
\uc{} is designed to run alongside them.

\subsection{LLM serving}
vLLM~\cite{kwon2023vllm} combines paged KV-cache management with
iteration-level (continuous) batching~\cite{yu2022orca}; SGLang~\cite{zheng2024sglang}
follows a similar execution model. Their extension points (e.g., logits
processors) see only the final logits, not intermediate hidden states, and
offer no mechanism for a per-request probe to observe the forward pass. Making
activation capture correct inside such an engine requires recovering, for every
row of every batched tensor, which request and which token position it belongs
to.

\section{Design Goals}
\label{sec:goals}

\uc{} was designed around five goals derived from running probes on live
traffic.

\textbf{G1: Configuration, not surgery.} What to capture should be a
reviewable artefact, separate from probe code and from the serving stack.

\textbf{G2: Act before the next token.} A probe that detects a problem should
be able to stop the request at the earliest point the engine allows.

\textbf{G3: Observation must not cost latency.} Monitoring probes must run off
the generation path, and a slow probe must degrade by losing data, not by
slowing generation, unless the operator explicitly chooses otherwise.

\textbf{G4: Isolation and ordering.} Probe state must never leak between
concurrent requests, and a stateful probe must see its activations in token
order.

\textbf{G5: Fail loudly, degrade predictably.} Misconfiguration and engine
incompatibilities must raise errors rather than produce silently wrong
activations; probe failures and timeouts must have explicit, bounded effects.

\section{Architecture}
\label{sec:arch}

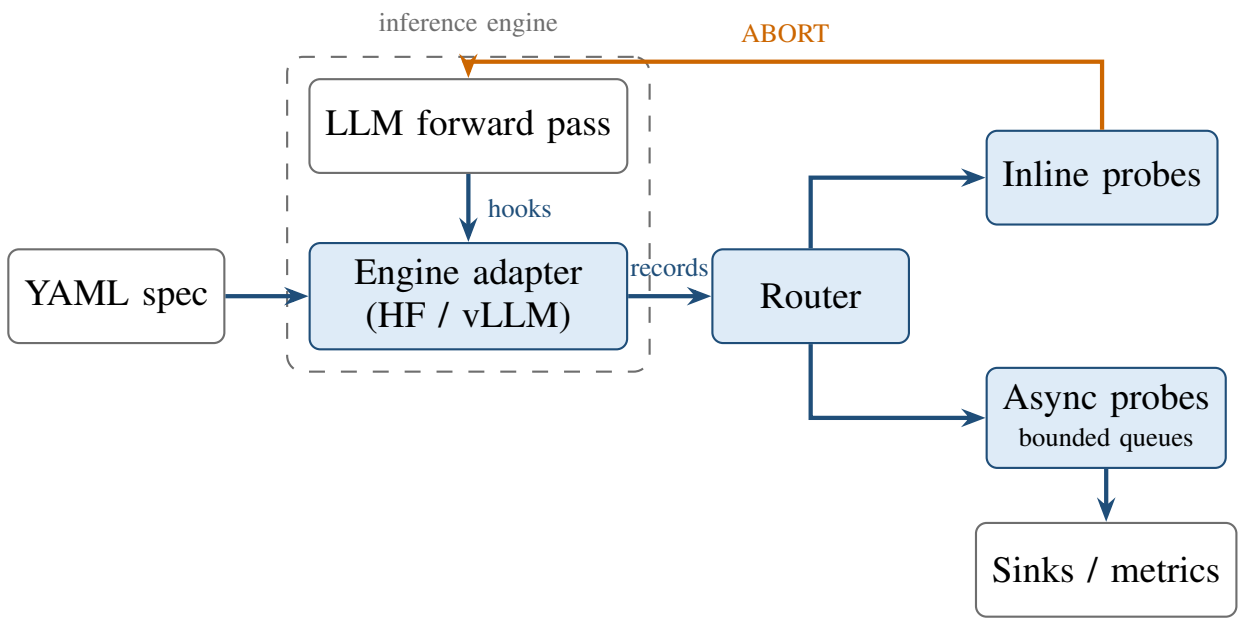
\begin{figure}[t]
\centering
\begin{tikzpicture}[
  font=\scriptsize,
  box/.style={draw=ucblue, fill=uclight, rounded corners=2pt, minimum height=0.62cm, align=center, inner sep=3pt},
  plain/.style={draw=ucgray, rounded corners=2pt, minimum height=0.62cm, align=center, inner sep=3pt},
  arr/.style={-{Stealth[length=1.6mm]}, thick, ucblue},
  darr/.style={-{Stealth[length=1.6mm]}, thick, ucorange},
  node distance=0.35cm and 0.45cm
]
  \node[plain] (spec) {YAML spec};
  \node[box, right=0.55cm of spec, minimum width=2.1cm] (adapter) {Engine adapter\\(HF / vLLM)};
  \node[plain, above=0.45cm of adapter, minimum width=2.1cm] (model) {LLM forward pass};
  \node[box, right=0.55cm of adapter, minimum width=1.3cm] (router) {Router};
  \node[box, above right=0.15cm and 0.5cm of router, minimum width=1.5cm] (inline) {Inline probes};
  \node[box, below right=0.15cm and 0.5cm of router, minimum width=1.5cm] (async) {Async probes\\[-1pt]{\tiny bounded queues}};
  \node[plain, below=0.35cm of async, minimum width=1.5cm] (sinks) {Sinks / metrics};

  \begin{scope}[on background layer]
    \node[draw=ucgray, dashed, rounded corners=3pt, fit=(model)(adapter), inner sep=4pt, label={[ucgray,font=\tiny]above:inference engine}] {};
  \end{scope}

  \draw[arr] (spec) -- (adapter);
  \draw[arr] (model) -- node[right, font=\tiny]{hooks} (adapter);
  \draw[arr] (adapter) -- node[above, font=\tiny]{records} (router);
  \draw[arr] (router) |- (inline);
  \draw[arr] (router) |- (async);
  \draw[arr] (async) -- (sinks);
  \draw[darr] (inline.north) -- ++(0,0.45) -| node[pos=0.25, above, font=\tiny, text=ucorange]{ABORT} (model.north);
\end{tikzpicture}
\caption{Data flow in \uc{}. An engine adapter installs forward hooks for the
extraction points in the spec, builds an \code{ActivationRecord} for every
matching (token, layer), and routes it. Inline probes return signals on the
generation path and may abort; async probes observe through bounded queues and
report to sinks.}
\label{fig:arch}
\end{figure}

\uc{} is a single Python package of six layered subpackages, each depending
only on those below it (Fig.~\ref{fig:arch}). \code{spec} defines the
extraction-point contract and the runtime \code{ActivationRecord}, with no
engine imports. \code{core} defines the probe interface. \code{router}
dispatches records to per-request probe instances and owns queues, worker
threads, intervention policies and metrics. \code{sinks} provides out-of-band
logging. \code{adapters} implements the engine-agnostic \code{EngineAdapter}
contract for Hugging Face and vLLM. Finally, \code{model} exposes
\code{ProbedModel}, a high-level front door that wires everything together.
Because the router only consumes \code{ActivationRecord}s, the entire pipeline
can be driven with synthetic data, which is how most of the test suite works.

Listing~\ref{lst:quickstart} shows the complete user-facing workflow: a
three-line function probe attached to GPT-2~\cite{radford2019gpt2} running on
a CPU.

\begin{lstlisting}[style=py, caption={A function probe attached to GPT-2 through \code{ProbedModel}.}, label={lst:quickstart}, float=t]
from undercurrent import ProbedModel, probe

SPEC = """
version: "1"
extraction_points:
  - name: last_prompt_token
    layers: 6
    tensor_type: residual_stream
    position: "prompt[-1]"
    probe_type: norm_gate
    probe_kind: single_shot
    execution_mode: inline
"""

@probe("norm_gate", threshold=500.0)
def norm_gate(record) -> float:
    return float(record.tensor.norm())  # abort if >= 500

with ProbedModel.from_pretrained("openai-community/gpt2",
        spec=SPEC, probes={"norm_gate": norm_gate}) as m:
    out = m.generate("The quick brown fox",
                     max_new_tokens=20)
result = out.probe_results["last_prompt_token"]
print(out.aborted, result.verdict)
\end{lstlisting}

\section{Extraction-Point Specification}
\label{sec:spec}

An \emph{extraction point} names a probe and the activations it receives
(Listing~\ref{lst:spec}). It selects one or more \textbf{layers}, a
\textbf{tensor type} (\code{residual\_stream}, \code{attn\_out}, \code{mlp\_out}
or \code{final\_norm}), a \textbf{position}, a \textbf{probe kind}
(\code{single\_shot} or \code{trajectory}) and an \textbf{execution mode}.
Specs are validated by pydantic models, published as a JSON Schema for editor
support, round-trip back to YAML, and can be checked offline with the
\code{undercurrent validate} command. \code{undercurrent inspect-model} lists a
model's hookable layers and supported tensor types.

\begin{lstlisting}[style=yaml, caption={A spec combining an inline guard with an asynchronous monitor.}, label={lst:spec}, float=t]
version: "1"
extraction_points:
  - name: guard              # can stop the request
    layers: 16
    tensor_type: residual_stream
    position: "prompt[-1]"
    probe_type: toxicity_linear
    probe_kind: single_shot
    execution_mode: inline
    intervention:
      mode: block_until_signal
      timeout_ms: 250
      on_timeout: abort      # fail closed
  - name: monitor            # observes, never blocks
    layers: [4, 8, 12]
    tensor_type: mlp_out
    position: "generated[*]"
    stride: 2
    until: generation_end
    probe_type: trajectory_probe
    probe_kind: trajectory
    execution_mode: async
    queue_depth: 8
\end{lstlisting}

\textbf{Position selectors.} A small grammar (Table~\ref{tab:positions})
addresses tokens relative to the prompt/generation boundary, which is what
probes are usually trained against: the last prompt token (\code{prompt[-1]}),
the first generated token (\code{generated[0]}, equivalently
\code{prompt[-1]+1}), or a continuous run of generated tokens with an optional
\code{stride} and an \code{until} condition (\code{generation\_end},
\code{fixed\_count} or \code{stop\_token}). Selectors are parsed once into
frozen values whose \code{matches()} method adapters call per token; malformed
selectors fail at parse time with the list of accepted forms (G5).

\begin{table}[t]
\caption{Position-selector grammar (0-based indices).}
\label{tab:positions}
\centering
\footnotesize
\begin{tabular}{@{}lll@{}}
\toprule
Form & Example & Matches \\
\midrule
absolute & \code{5} & token 5 overall \\
prompt index & \code{prompt[-1]} & last prompt token \\
generated index & \code{generated[0]} & first generated token \\
offset & \code{prompt[-1]+1} & shifted after resolution \\
all generated$^\dagger$ & \code{generated[*]} & every generated token \\
open slice$^\dagger$ & \code{generated[5:]} & generated $i \geq 5$ \\
closed slice$^\dagger$ & \code{generated[2:8]} & generated $2 \le i < 8$ \\
\bottomrule
\multicolumn{3}{@{}l}{\scriptsize $^\dagger$Continuous: accepts \code{stride} and \code{until}.}
\end{tabular}
\end{table}

\section{Probes and Execution Model}
\label{sec:exec}

\subsection{Probe lifecycle and isolation}
A probe class implements three methods. For each request and extraction point
the router calls \code{spawn} to create a fresh instance, \code{on\_start}
once, \code{on\_activation} once per matching record (returning an optional
\code{ProbeSignal} with action \code{continue}, \code{flag} or \code{abort}),
and \code{on\_end} exactly once, even if generation was aborted or the point
never matched, returning a \code{ProbeResult} verdict. Stateless checks can be
written as plain functions with the \code{@probe} decorator; a returned float
is compared against a threshold. Probes can be registered by name and
distributed by third-party packages through the \code{undercurrent.probes}
entry-point group.

Each instance is keyed by its request id and extraction-point name and is
never shared across requests, extraction points or threads, so per-request state on
\code{self} needs no locks (G4). The library enforces this: defining a mutable
class attribute (\code{list}, \code{dict}, \code{set}, \code{bytearray}) on a
probe subclass raises \code{TypeError} at class-definition time, eliminating a
common source of cross-request leakage.

\subsection{Inline execution}
An inline extraction point calls the probe synchronously inside
\code{router.route()} on the adapter's thread and returns its signal to the
adapter, which stops the request on \code{abort} (G2). If an inline probe
spends $t_p$ seconds per activation and matches $m$ (layer, token) pairs per
decode step, its added latency per step is approximately
\begin{equation}
  \Delta_{\text{inline}} \approx \sum_{e \in \mathcal{E}_{\text{inline}}} m_e \left( t_c + t_{p,e} \right),
  \label{eq:inline}
\end{equation}
where $t_c$ is the capture cost (slicing the row and copying it to the host).
This makes the cost of inline probing explicit and attributable: it scales
with the number of matched positions and with probe compute, not with model
size.

\textbf{Intervention policies.} By default (\code{mode: reject}) the router
waits as long as the probe takes. With \code{mode: block\_until\_signal} the
router runs the probe on its worker pool and waits at most
\code{timeout\_ms}; if the probe times out or raises, it substitutes a fallback
signal chosen by \code{on\_timeout}: \code{continue} fails \emph{open}
(appropriate for quality monitors), and \code{abort} fails \emph{closed}
(appropriate for safety gates where unchecked output is worse than truncated
output). The per-step cost of a bounded point is then capped at
$\min(t_{p}, \tau)$ for timeout $\tau$.

\textbf{Circuit breaker.} A probe that keeps timing out would cost $\tau$ on
every matching token. Following the circuit-breaker pattern~\cite{nygard2018releaseit},
the router counts consecutive timeouts and exceptions per extraction point
across requests; after \code{circuit\_breaker\_threshold} (default 5) it
downgrades the point to plain \code{reject} mode, logs a warning, and reports a
\code{flag} signal to the attached sink (G5). Any success resets the count.

\subsection{Asynchronous execution}
An async extraction point never blocks generation (G3). On
\code{register\_request} the router creates, for each async point, a
\emph{binding}: a fresh probe instance, a bounded queue of \code{queue\_depth}
entries (default 32), and a single drain loop scheduled on a shared thread
pool (default size $\min(32, 4 \times \text{cores})$). \code{route()} only
enqueues and returns immediately. Because each queue has exactly one consumer,
a trajectory probe observes its activations strictly in routing order, on one
thread at a time (G4). When a request ends, \code{end\_request} closes the
queue, drains what remains (bounded by a 30\,s default \code{drain\_timeout}),
and only then calls \code{on\_end}, so an async verdict covers every
activation that entered the queue, even after an inline abort.

\textbf{Overflow policies.} When a binding's queue is full the router applies
one of three policies (Table~\ref{tab:overflow}). The two drop policies keep
the generation path wait-free and convert overload into counted data loss; the
\code{block} policy preserves every activation at the cost of slowing
generation to the probe's pace, which suits offline evaluation and data
collection. Per-binding queue depth, drops, activation latency and probe errors
are exported to a pluggable metrics sink.

\begin{table}[t]
\caption{Overflow policies for asynchronous bindings.}
\label{tab:overflow}
\centering
\footnotesize
\begin{tabular}{@{}>{\raggedright\arraybackslash}p{1.85cm}>{\raggedright\arraybackslash}p{2.55cm}>{\raggedright\arraybackslash}p{3.0cm}@{}}
\toprule
Policy & On full queue & Trade-off \\
\midrule
\code{drop\_oldest} (default) & evict oldest, accept new & probe sees recent window; no stall \\
\code{drop\_newest} & reject incoming & probe sees earliest prefix; no stall \\
\code{block} & \code{route()} waits & lossless; generation slows \\
\bottomrule
\end{tabular}
\end{table}

\textbf{Observation sinks.} Async signals and results flow to a sink: an
NDJSON \code{FileLogSink}, or a \code{WebhookLogSink} that posts from a
background thread with exponential-backoff retries, a bounded send queue and
dead-lettering to a local file. Tensors are summarised by shape and dtype rather
than serialised, and composable redaction filters (\code{redact\_keys},
\code{drop\_keys}, \code{chain}) control what leaves the process; the webhook
sink redacts prompt text by default.

\section{Engine Adapters}
\label{sec:adapters}

Every adapter implements the four methods of the \code{EngineAdapter} base
class: loading a model, registering and unregistering a request's extraction
points, and generating.
The contract requires one \code{ActivationRecord} per matching (token, layer)
and that an inline abort stop generation as promptly as the engine allows.

\subsection{Hugging Face Transformers}
The HF adapter~\cite{wolf2020transformers} registers PyTorch~\cite{paszke2019pytorch}
forward hooks on the decoder layers (for \code{residual\_stream}) and on their
attention and MLP submodules. Hooks fire inside \code{model.forward()}, on the
same thread that HF's decode loop later uses to evaluate stopping criteria.
Each matching hook calls \code{router.route()}; an inline \code{abort} sets a
flag that a custom \code{StoppingCriteria} reads before the next token is
produced. Because the timeout of a \code{block\_until\_signal} point is
enforced inside \code{route()}, every extraction point has already had its
bounded chance to intervene by the time the stopping criterion runs, so the
added latency of a step is predictable by construction. The HF adapter
processes one sequence at a time, which gives it the cleanest abort semantics
and makes it the natural backend for research, evaluation and CPU deployment.

\subsection{vLLM}
\label{sec:vllm}

\begin{algorithm}[t]
\caption{Row-to-position mapping for one scheduler step}
\label{alg:seqmap}
\small
\begin{algorithmic}[1]
\Require Step metadata: scheduled requests $r_1..r_k$ in flattening order, with
scheduled token counts $n_j$ and tokens already computed $c_j$; per-request
prompt length $P[r]$ and generated count $G[r]$
\Ensure One $(\text{row}, r, \text{pos}, \text{isGen}, \text{genIdx})$ per row
\State $\text{offset} \gets 0$
\For{$j \gets 1$ \textbf{to} $k$}
  \If{$r_j$ not registered} \textbf{raise} \textsc{SeqMapperError} \EndIf
  \For{$\ell \gets 0$ \textbf{to} $n_j - 1$}
    \State $\text{pos} \gets c_j + \ell$; \quad $\text{isGen} \gets \text{pos} \ge P[r_j]$
    \State \textbf{emit} $(\text{offset}+\ell,\ r_j,\ \text{pos},\ \text{isGen},\ \text{pos}-P[r_j])$
    \If{isGen} $G[r_j] \gets G[r_j] + 1$ \EndIf
  \EndFor
  \State $\text{offset} \gets \text{offset} + n_j$
\EndFor
\end{algorithmic}
\end{algorithm}

\textbf{The attribution problem.} Under continuous batching each decoder layer
processes a single flattened tensor of shape $[T, d]$, where the $T$ rows
concatenate prefill chunks and single decode tokens from whichever requests the
scheduler selected for this step. Requests join and leave independently, row
slots are reused, and long prompts are split across several prefill steps.
There is therefore no stable ``token position $N$''; the mapping must be
recomputed every step from scheduler metadata.

\textbf{Worker extension.} \uc{} mixes a probing worker extension into
each vLLM worker, using the engine's \code{worker\_extension\_cls} argument.
The extension wraps the model runner's \code{execute\_model} to retain the
current step's \code{scheduler\_output}, and lazily installs forward hooks on
the decoder layers, attention and MLP blocks, and the final norm. The first
hook to fire in a step reads the persistent batch layout (request ids in row
order, tokens already computed, tokens scheduled this step) into a
\code{StepBatchMetadata}; a pure-Python \code{SeqIdMapper} then translates each
row into a request id and absolute token position (Algorithm~\ref{alg:seqmap}).
Later hooks in the same step reuse the mapping. Rows that match a registered
extraction point are copied to the host as \code{float32} and routed; hooks
with no interested extraction point return immediately. The mapper is
independent of torch and vLLM and is unit-tested against hand-built step
layouts covering single-shot and chunked prefill, decode, co-scheduled requests,
reused row slots, and requests that skip a step.

\textbf{Process topology.} Inline decisions require \code{route()}'s answer,
but vLLM's \code{collective\_rpc} only lets the driver call into workers. When
the worker shares the driver's process, the adapter binds the live router into
the worker and routes synchronously from the hook. When the engine core runs in
a subprocess, the router cannot be serialised across the boundary, so the
adapter falls back to \emph{polling}: workers buffer records, and after every
engine output the driver pops them over RPC and routes them in capture order,
off the event loop. Capture and mapping are identical on both paths; only the
location of \code{route()} differs, at the cost of up to one step of latency on
the polling path.

\textbf{Aborts.} An inline abort is delivered through the engine's own
\code{abort(request\_id)}. Because vLLM schedules a whole step before checking
for aborted requests, the earliest a request can stop is the next scheduler
iteration; at least one extra token past the triggering step should be expected,
and a few under heavy load. For the same reason, a \code{block\_until\_signal}
wait on vLLM stalls every request in the batch, not only the probed one;
the adapter warns about this, and the documentation recommends \code{reject}
mode or async execution under concurrency.

\textbf{Fused residuals.} vLLM's Llama-family decoder layers defer the residual
addition into the next layer's fused normalisation and return a
\code{(hidden\_states, residual)} pair in which \code{hidden\_states} is the
MLP output. Capturing the first element, as a naive hook would, yields the MLP
output rather than the residual stream. \uc{} computes \code{hidden\_states +
residual} out of place in the model's dtype (vLLM's fused kernel later
overwrites both tensors in place), passes single-tensor outputs (e.g., GPT-2)
through unchanged, and raises \code{VLLMAdapterLimitationError} on any other
output shape rather than guessing (G5).

\textbf{Defensive coupling.} The adapter reads undocumented vLLM internals.
It therefore pins a tested vLLM minor version and checks it at runtime, verifies
the introspected structures on the first step, forces the V1 model runner and
stable request ids (each with an explicit opt-out), and refuses multi-worker
topologies unless the user overrides the check. With the override, results
from tensor-parallel workers are merged and duplicate events deduplicated.

\section{Implementation}
\label{sec:impl}

\uc{} 0.1.0 is about 10.8k lines of typed Python (it ships \code{py.typed})
supporting Python 3.10--3.13, PyTorch 2.1+ and Transformers 4.40--5.x; vLLM is
an optional dependency, never installed by default, with a supported range of
\code{>=0.28,<0.29}. The test suite (about 13.6k lines, roughly 700 test
functions) runs on CPU in continuous integration, including a job pinned to the
declared dependency floors, and exercises the vLLM worker extension against a
fake model runner. GPU tests are run before each release by a gate script;
0.1.0 passed it on an NVIDIA L4 with vLLM 0.28.0. Errors raised deliberately
derive from a common \code{ProbingError} hierarchy and state how to fix the
problem. The public API, deprecation policy and experimental surface (the vLLM
adapter is marked experimental) are documented. The repository also contains a
reference OpenAI-compatible wire format and HTTP server that returns probe
verdicts alongside completions, a content-safety wiring demo, Colab notebooks,
and a probe-training example (Sec.~\ref{sec:case}).

\section{Evaluation}
\label{sec:eval}

We report three kinds of evidence: correctness of activation capture across
backends, overhead measurements from the library's benchmark harness, and an
end-to-end case study that trains a probe and deploys it inline.

\subsection{Cross-backend capture equivalence}
Activation capture is only useful if it reads the tensor the probe was trained
on. The release gate includes a GPU test that builds a tiny random Llama
(fused residual) and a tiny random GPT-2 (single-tensor output), loads both in
\code{float32} under the HF and vLLM backends, generates greedily from the same
prompt, and requires the \code{residual\_stream} captures at the first and last
layer to match token by token with relative error at most $10^{-3}$. The error
this test guards against, capturing the MLP output instead of the residual
stream, has relative error of order one. \uc{} 0.1.0 passed this test on an
NVIDIA L4 (vLLM 0.28.0, PyTorch 2.13.0, CUDA 13.0), as did the vLLM integration
tests covering single-shot capture, \code{generated[*]} trajectories with a
concurrent second request, and inline aborts.

\subsection{Probing overhead}
\label{sec:overhead}

\textbf{Setup.} We ran the repository's benchmark harness
(\code{benchmarks/run.py}) on the Hugging Face backend with GPT-2 (124M
parameters, 12 layers) on a laptop CPU (AMD Ryzen 7 6800H, 16 logical cores,
Linux under WSL2; Python 3.14, PyTorch 2.14 CPU build, Transformers 5.18,
\uc{} 0.1.0 at a clean commit). Each scenario generated exactly 64 tokens
greedily for each of 16 prompts of 64 tokens, after 2 warm-up batches, and was
repeated 3 times; every scenario saw byte-identical prompts. The
\code{baseline} scenario runs plain \code{model.generate()} with no \uc{} code;
\code{probed-0} wraps the model in \code{ProbedModel} with an empty spec;
\code{inline-}$N$ and \code{async-}$N$ attach $N$ extraction points spread
evenly across layers, each capturing \code{residual\_stream} at every
generated token (\code{generated[*]}) for a trajectory probe. The \emph{norm}
probe computes an L2 norm (one reduction and a host sync); the \emph{mlp} probe
is a random-weight $768\!\to\!256\!\to\!1$ head, roughly the cost of a trained
classifier. \code{mixed} runs four inline and four async points. No probe
aborts, so all scenarios do identical model work.

\begin{table}[t]
\caption{Probing overhead, HF backend, GPT-2 on CPU (16 prompts $\times$ 64
generated tokens, 3 repeats). Throughput is mean $\pm$ s.d.\ over repeats;
TTLT is the median time to last token per request; ``inline cost'' is the time
spent inside inline probes per generated token.}
\label{tab:overhead}
\centering
\footnotesize
\setlength{\tabcolsep}{4pt}
\begin{tabular}{@{}lrrrr@{}}
\toprule
Scenario & Tok/s & TTLT (s) & $\Delta$ tok/s & Inline cost \\
\midrule
\code{baseline}      & $21.3 \pm 1.1$ & 2.96 & ---     & --- \\
\code{probed-0}      & $21.9 \pm 0.1$ & 2.84 & $+2.9\%$ & --- \\
\code{inline-1}      & $22.3 \pm 0.5$ & 2.84 & $+4.7\%$ & 0.03\,ms \\
\code{async-1}       & $21.1 \pm 0.9$ & 2.99 & $-1.0\%$ & --- \\
\code{inline-8}      & $22.7 \pm 1.3$ & 2.76 & $+6.4\%$ & 0.19\,ms \\
\code{async-8}       & $22.3 \pm 0.1$ & 2.86 & $+4.6\%$ & --- \\
\code{inline-8-mlp}  & $20.5 \pm 1.6$ & 3.13 & $-3.7\%$ & 1.28\,ms \\
\code{async-8-mlp}   & $21.6 \pm 0.5$ & 2.91 & $+1.4\%$ & --- \\
\code{mixed} (4+4)   & $22.3 \pm 0.2$ & 2.85 & $+4.6\%$ & 0.39\,ms \\
\bottomrule
\end{tabular}
\end{table}

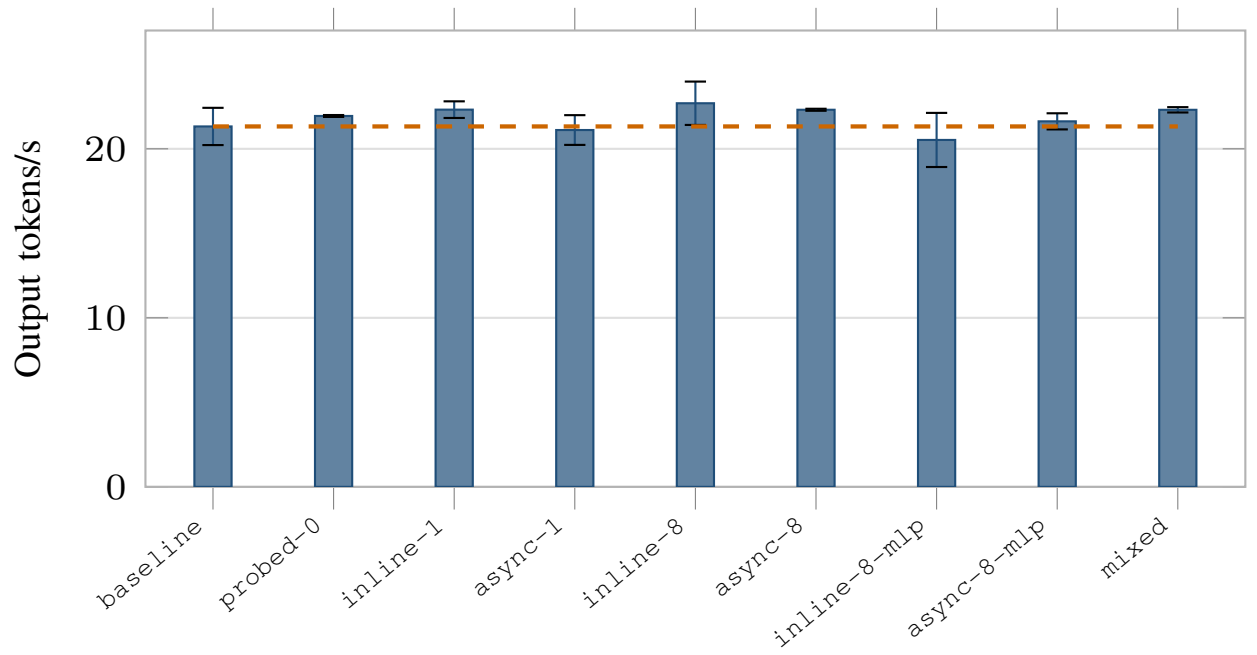
\begin{figure}[t]
\centering
\begin{tikzpicture}
\begin{axis}[
  width=\columnwidth, height=4.6cm,
  ybar, bar width=7pt,
  ymin=0, ymax=27,
  ylabel={Output tokens/s}, ylabel style={font=\scriptsize},
  symbolic x coords={baseline,probed-0,inline-1,async-1,inline-8,async-8,inline-8-mlp,async-8-mlp,mixed},
  xtick=data,
  xticklabel style={rotate=40, anchor=east, font=\tiny\ttfamily},
  yticklabel style={font=\scriptsize},
  ymajorgrids, grid style={gray!25},
  error bars/y dir=both, error bars/y explicit,
  enlarge x limits=0.07,
  axis line style={gray!60},
]
\addplot[fill=ucblue!70, draw=ucblue] coordinates {
  (baseline,21.324) +- (0,1.102)
  (probed-0,21.943) +- (0,0.056)
  (inline-1,22.318) +- (0,0.492)
  (async-1,21.112) +- (0,0.880)
  (inline-8,22.696) +- (0,1.283)
  (async-8,22.311) +- (0,0.060)
  (inline-8-mlp,20.525) +- (0,1.601)
  (async-8-mlp,21.626) +- (0,0.477)
  (mixed,22.308) +- (0,0.158)
};
\draw[ucorange, dashed, thick] (axis cs:baseline,21.324) -- (axis cs:mixed,21.324);
\end{axis}
\end{tikzpicture}
\caption{Generation throughput per scenario (error bars: $\pm$1 s.d.\ over
three repeats). The dashed line is the unprobed baseline. All probed
scenarios fall within the baseline's run-to-run variation.}
\label{fig:throughput}
\end{figure}

\textbf{Results.} Table~\ref{tab:overhead} and Fig.~\ref{fig:throughput}
summarise the runs. On this configuration the overhead of the probing
machinery is below the measurement noise: every probed scenario is within
$\pm 7\%$ of the baseline's throughput, whose own standard deviation across
repeats is $5\%$, and several probed scenarios are nominally \emph{faster} than
the baseline, which ran first. We therefore do not claim a specific overhead
figure from these runs; we claim only that it is not distinguishable from
variance here. The direct measurement of inline probe time is more informative
and agrees with Eq.~\eqref{eq:inline}. The norm probe costs about
$24\,\mu$s per activation and the MLP head about $160\,\mu$s, so eight inline
MLP probes add $1.28$\,ms per generated token, about $2.7\%$ of the
$\approx 47$\,ms decode step, and it is the one inline scenario whose median
TTLT rises above the baseline. The same eight MLP probes run asynchronously
spend $2.0$\,s of probe compute per repeat off the generation path, with the
async queues never holding more than one pending activation.

These are CPU measurements on a small model, where a decode step is slow
relative to probe compute. On a GPU, decode steps are much shorter, so the
same inline probe compute, and the device-to-host copy that precedes it, is a
larger fraction of each step. The harness supports vLLM on GPUs with the same
scenarios and output schema, and the documentation specifies a reference
matrix (Llama-3.1-8B on an A100/H100); we have not yet run it, and we make no
claims about GPU overhead here.

\textbf{Backpressure.} The \code{async-slow} scenario attaches one async probe
that sleeps $100$\,ms per activation, about twice the decode step, with a queue
depth of 4, once per overflow policy (Table~\ref{tab:backpressure}). With
either drop policy, token generation was unaffected (median TTLT $2.67$--$2.69$\,s
against the baseline's $2.96$\,s) and the probe processed roughly half of the
1008 activations per repeat, every loss counted in the metrics. End-to-end
latency exceeded TTLT by about $0.44$\,s, the end-of-request drain of at most
$4 \times 100$\,ms of queued work, as designed. With \code{block}, nothing was
dropped but generation slowed to the probe's pace (TTLT $5.96$\,s, $-53\%$
throughput). Each policy behaved exactly as its contract states (G3).

\begin{table}[t]
\caption{Backpressure with a slow async probe (100\,ms per activation, queue
depth 4). Activations and drops are per repeat (1008 activations offered).}
\label{tab:backpressure}
\centering
\footnotesize
\setlength{\tabcolsep}{3.5pt}
\begin{tabular}{@{}lrrrrr@{}}
\toprule
Policy & TTLT (s) & Latency (s) & Tok/s & Processed & Dropped \\
\midrule
baseline (no probe)    & 2.96 & 2.96 & 21.3 & --- & --- \\
\code{drop\_oldest}    & 2.67 & 3.11 & 20.6 & 479 & 529 \\
\code{drop\_newest}    & 2.69 & 3.11 & 19.7 & 503 & 505 \\
\code{block}           & 5.96 & 6.44 & 9.9  & 1008 & 0 \\
\bottomrule
\end{tabular}
\end{table}

\subsection{Case study: from trained probe to inline guard}
\label{sec:case}

To exercise the full workflow a practitioner would follow, we used the
repository's \code{examples/train\_probe} pipeline, which (1) collects one
activation per labelled prompt through \code{ProbedModel} with a collector
probe, (2) trains a probe in plain PyTorch, (3) evaluates it on held-out data,
(4) saves it as \code{safetensors} plus a JSON descriptor of the layer, tensor
type and position it was trained on, and (5) reloads it as an inline
\uc{} probe whose spec is generated from that descriptor, so the deployed probe
reads exactly the tensor it was trained on.

\textbf{Setup.} We sampled 2{,}000 class-balanced comments from the Civil
Comments training split~\cite{borkan2019civil} (labelled toxic if the crowd
toxicity score is at least 0.5), split them 70/15/15 per class, and trained a
logistic-regression probe on GPT-2's residual stream at the last prompt token
(\code{prompt[-1]}) for each of five layers, keeping the epoch with the lowest
validation loss. Each run took about four minutes on the same laptop CPU,
almost all of it spent collecting activations.

\begin{table}[t]
\caption{Linear toxicity probes on GPT-2's residual stream at the last prompt
token (Civil Comments, 1{,}400 train / 300 val / 300 test, balanced).}
\label{tab:probe}
\centering
\footnotesize
\begin{tabular}{@{}lccccc@{}}
\toprule
Layer & 0 & 3 & 6 & 9 & 11 \\
\midrule
Val AUROC   & 0.794 & 0.814 & 0.820 & 0.821 & 0.829 \\
Test AUROC  & 0.816 & 0.846 & 0.838 & 0.847 & 0.854 \\
Test acc.   & 0.740 & 0.777 & 0.777 & 0.767 & 0.783 \\
Test F1     & 0.735 & 0.782 & 0.777 & 0.768 & 0.787 \\
\bottomrule
\end{tabular}
\end{table}

\textbf{Results.} Table~\ref{tab:probe} reports the held-out metrics. A linear
read-out of a single residual-stream vector separates toxic from non-toxic
comments with test AUROC between 0.82 and 0.85. The embedding-adjacent layer 0
is the weakest; layers 3 to 11 are within about 0.015 AUROC of one another,
which is inside the sampling error of a 300-example test set (roughly $\pm 0.02$),
so these runs do not identify a best layer. Loaded back as an inline probe at
layer 6 with a 0.5 threshold, the probe scored the benign prompt
\emph{``Thanks for sharing, the recipe looks great.''} at 0.025 and let
generation proceed, and scored the insult \emph{``Shut up, you are a worthless
idiot.''} at 0.991 and aborted the request at the first stopping check: the
verdict was reached during prefill, and generation stopped after a single
token (a newline) of the 20 allowed, returning the probe's verdict together
with the abort reason.

We stress what this case study does and does not show. It demonstrates that
\uc{} carries a probe from training to an inline abort with no glue code beyond
the example's 20-line loader, and that the deployed probe reads the same
activation it was trained on. It is not a toxicity detector: GPT-2 is small,
the sample is modest, Civil Comments labels carry rater biases, and we have
not evaluated robustness or calibration. Probe quality is the user's
responsibility (Sec.~\ref{sec:limits}).

\section{Comparison with Related Tools}
\label{sec:comparison}

Table~\ref{tab:comparison} positions \uc{} against representative tools along
the dimensions that matter for production probing. The interpretability
libraries are far more expressive for research: they patch, edit and attribute,
whereas \uc{} only reads. The guardrail tools are model-agnostic and come with
trained policies, whereas \uc{} requires access to the forward pass and ships
no trained probes. \uc{}'s distinguishing combination is in-engine capture
under continuous batching, a request-level abort from a probe verdict,
latency-free observation with explicit backpressure, enforced per-request
isolation, and a declarative, schema-validated configuration. In practice we
expect \uc{} to be used together with these tools: probes are trained with a
research library or plain PyTorch and deployed with \uc{}, while a text-level
classifier provides an independent check on the final output.

\begin{table*}[t]
\caption{Comparison with related tools (reviewed October 2026). \checkmark{} supported; $\circ$ partial; --- not supported or not a goal; n/a not applicable.}
\label{tab:comparison}
\centering
\footnotesize
\setlength{\tabcolsep}{3pt}
\begin{tabular}{@{}lllccccc@{}}
\toprule
Tool & Purpose & Operates on & In vLLM serving & Abort from probe & Async observation & Declarative spec & License \\
\midrule
\uc{} 0.1 & Production probing & activations & $\circ^{a}$ & \checkmark & \checkmark & \checkmark{} (YAML) & Apache-2.0 \\
TransformerLens~\cite{nanda2022transformerlens} & Interpretability research & activations & --- & $\circ^{b}$ & --- & --- & MIT \\
NNsight~\cite{fiottokaufman2025nnsight} & Interpretability research & activations & \checkmark & $\circ^{b}$ & $\circ$ & --- & MIT \\
pyvene~\cite{wu2024pyvene} & Interventions & activations & --- & $\circ^{b}$ & --- & $\circ$ (dict) & Apache-2.0 \\
baukit~\cite{bau2022baukit} & Research utilities & activations & --- & $\circ^{b}$ & --- & --- & MIT \\
Llama Guard~\cite{inan2023llamaguard} & Safety classifier & text & separate model & n/a & n/a & n/a & Llama Comm. \\
NeMo Guardrails~\cite{rebedea2023nemo} & Programmable rails & text & n/a & $\circ^{c}$ & --- & \checkmark{} (Colang) & Apache-2.0 \\
\bottomrule
\multicolumn{8}{@{}p{0.97\textwidth}@{}}{\scriptsize $^{a}$Validated with one model worker; multi-worker topologies require an explicit override. $^{b}$Activations can be read and edited during generation, but stopping a single request from a probe verdict requires a custom generation loop. $^{c}$Streaming output rails end the stream when a text chunk is blocked.}
\end{tabular}
\end{table*}

\section{Limitations and Future Work}
\label{sec:limits}

\textbf{Probe quality.} \uc{} is machinery, not a safety product. A probe's
precision and recall hold only on data like its evaluation data; probes can
learn shortcuts, degrade under distribution shift (new checkpoints, prompt
templates or domains), and their adversarial robustness is unknown. A probe
should never be the sole safety control.

\textbf{vLLM coverage.} The vLLM adapter is experimental. It supports one vLLM
minor version, the V1 model runner and eager execution (CUDA-graph replay
bypasses Python hooks); it is validated with a single model worker; KV-cache
extraction is rejected; and negative generated-token selectors such as
\code{generated[-1]} never match during streaming because the final length is
unknown. A bare \code{vllm serve} process loads the extension but does not
drive capture; HTTP serving currently requires embedding the adapter, as the
reference server does. There is no GPU continuous integration yet.

\textbf{HF throughput.} The HF adapter handles one sequence at a time without
beam search, which suits research and low-volume serving but not high-throughput
production, where the vLLM adapter is intended.

\textbf{Interventions.} The only intervention in 0.1 is abort; probes run on
the host after a device-to-host copy.

The roadmap includes a built-in OpenAI-compatible \code{undercurrent serve}
server, probe save/load and sharing via the Hugging Face Hub, validated tensor
and pipeline parallelism on vLLM, GPU-resident probes that avoid host copies,
activation-steering interventions, and adapters for SGLang, TGI and llama.cpp
through the engine-agnostic \code{EngineAdapter} interface.

\section{Responsible Use}
\label{sec:ethics}

Reading a model's internal state per request raises privacy considerations
distinct from text logging: activations and probe verdicts are derived from
user input. \uc{}'s sinks summarise tensors rather than storing them, the
webhook sink redacts prompt text by default, and redaction filters are
composable. The project documentation discourages covert surveillance and the
inference of sensitive personal traits, and recommends that operators disclose
what they monitor and retain it only as long as needed. The bundled
content-safety example uses random or dummy weights and is explicitly a wiring
demonstration, not a classifier.

\section{Conclusion}
\label{sec:conclusion}

Activation probes have shown that a model's internal state often reveals what
its output will be before that output exists. \uc{} makes such probes
deployable: a declarative spec says what to read, an execution model gives
every probe an explicit latency contract (inline with optional timeouts, or
asynchronous with bounded backpressure), and engine adapters make capture
correct inside both a simple Hugging Face decode loop and vLLM's continuously
batched scheduler. We hope that an open, engine-agnostic substrate for
production probing will make it easier to turn interpretability research into
runtime safeguards, and we welcome contributions, especially new engine
adapters.

\section*{Availability}
\uc{} is released under the Apache-2.0 license. Source code:
\url{https://github.com/wrynx/undercurrent}; documentation:
\url{https://wrynx.github.io/undercurrent/}; package:
\code{pip install undercurrent}.

\bibliographystyle{IEEEtran}
\bibliography{references}

\end{document}